\documentclass[10pt,twocolumn,letterpaper]{article}

\usepackage{iccv}              

\definecolor{iccvblue}{rgb}{0.21,0.49,0.74}
\usepackage[pagebackref,breaklinks,colorlinks,allcolors=iccvblue]{hyperref}
\usepackage{tikz}
\usepackage{makecell} 

\usepackage{algpseudocode}
\usepackage{amsmath}
\usepackage{amsfonts}
\usepackage{float}
\usepackage{booktabs} 
\usepackage{multirow} 
\usepackage[linesnumbered,ruled,vlined]{algorithm2e}

\def\paperID{14988} 
\def\confName{ICCV}
\def\confYear{2025}

\title{TriQDef: Disrupting Semantic and Gradient Alignment to Prevent Adversarial Patch Transferability in Quantized Neural Networks}

\author{Amira Guesmi\\
NYU Abu Dhabi\\
UAE\\
\and
Bassem Ouni\\
Technology Innovation Institute\\
UAE\\
\and
Muhammad Shafique\\
NYU Abu Dhabi\\
UAE\\
}

\begin{document}
\maketitle

\begin{abstract}
Quantized Neural Networks (QNNs) are increasingly deployed in edge and resource-constrained environments due to their efficiency in computation and memory usage. While shown to distort the gradient landscape and weaken conventional pixel-level attacks, it provides limited robustness against patch-based adversarial attacks—localized, high-saliency perturbations that remain surprisingly transferable across bit-widths. Existing defenses either overfit to fixed quantization settings or fail to address this cross-bit generalization vulnerability.
We introduce \textbf{TriQDef}, a tri-level quantization-aware defense framework designed to disrupt the transferability of patch-based adversarial attacks across QNNs. TriQDef consists of: (1) a \emph{Feature Disalignment Penalty (FDP)} that enforces semantic inconsistency by penalizing perceptual similarity in intermediate representations; (2) a \emph{Gradient Perceptual Dissonance Penalty (GPDP)} that explicitly misaligns input gradients across bit-widths by minimizing structural and directional agreement via Edge IoU and HOG Cosine metrics; and (3) a \emph{Joint Quantization-Aware Training Protocol} that unifies these penalties within a shared-weight training scheme across multiple quantization levels.
Extensive experiments on CIFAR-10 and ImageNet demonstrate that TriQDef reduces Attack Success Rates (ASR) by over 40\% on unseen patch and quantization combinations, while preserving high clean accuracy. Our findings underscore the importance of disrupting both semantic and perceptual gradient alignment to mitigate patch transferability in QNNs.
\end{abstract}

\section{Introduction}
\label{sec:intro}

Quantized Neural Networks (QNNs) offer a compelling trade-off by significantly reducing memory and compute requirements while maintaining competitive accuracy~\cite{liu2021flexi, katare2023survey, zhang2021medq, tonellotto2021neural, hernandez2024optimizing}. Prior studies have shown that quantization can distort gradient landscapes, thereby weakening the effectiveness of traditional pixel-level adversarial attacks~\cite{Li2024InvestigatingTI, yang2024quantization}. However, such gradient masking effects offer little protection against more structured threats.

\textit{Adversarial patch attacks}~\cite{lavan, googleap, chen2022shape} pose a unique challenge by inserting localized, high-saliency patterns that hijack model predictions. Unlike subtle pixel perturbations, these patches are robust to input variation and generalize across architectures and quantization levels. Crucially, our analysis reveals that even under aggressive quantization (e.g., 2-bit), adversarial patches crafted on full-precision models maintain high success rates—highlighting a critical blind spot in current quantization-aware defenses.


Existing defenses such as adversarial training~\cite{madry2017towards} or Patch-Based Adversarial Training (PBAT)~\cite{rao2020adversarial} either overfit to specific patch patterns or fail to disrupt these cross-bit generalization pathways. Input preprocessing-based defenses such as \cite{tarchoun2023jedi, nie2022DiffPure} impose significant computational overhead, undermining the efficiency gains that quantization aims to provide. Recent quantization-aware methods like DWQ~\cite{fu2021double} and feature denoising~\cite{song2020improving} overlook the perceptual and optimization-level alignment that underlies patch transferability.

In this work, we introduce \textbf{TriQDef}, a unified defense framework designed to explicitly disrupt the core enablers of patch-based adversarial transferability in QNNs. Our analysis reveals that quantized models—even under extreme bit-width reduction—exhibit surprisingly high vulnerability to transferable adversarial patches. This phenomenon arises from persistent alignment in both internal features and input gradient signals across bit-widths. TriQDef addresses this vulnerability through three synergistic components that target semantic and optimization-level consistency:

\noindent\textbf{Feature Disalignment Penalty (FDP)} enforces semantic divergence by penalizing perceptual similarity in feature maps across quantized variants. Using differentiable variants of Edge IoU and HOG Cosine similarity, FDP encourages each bit-width to develop unique feature representations, thus weakening patch generalization.

\noindent\textbf{Gradient Perceptual Dissonance Penalty (GPDP)} misaligns the saliency landscape by penalizing structural and perceptual gradient similarity across bit-widths. It directly targets gradient-level alignment that facilitates adversarial transfer, extending beyond cosine similarity to include perceptual alignment in edge structures and gradient orientations.

\noindent\textbf{Bit-Width-Aware Curriculum Training (BACT)} stabilizes learning across bit-widths by progressively introducing lower-bit models during training. This curriculum design improves convergence under quantization noise while jointly training a shared-weight model that integrates both FDP and GPDP constraints.

\noindent\textbf{Key Contributions:}
\begin{itemize}
    \item We conduct the first systematic study of patch-based transferability in QNNs, demonstrating that adversarial patches remain highly effective across quantization levels, including 2-bit regimes.
    \item We propose \textbf{TriQDef}, the first tri-component defense targeting both feature and gradient alignment, explicitly designed to prevent cross-bit patch generalization.
    \item We introduce perceptual alignment metrics—Edge IoU and HOG Cosine Similarity—as theoretically justified tools to quantify and disrupt semantic and gradient-level alignment across bit-widths. These metrics provide a principled alternative to cosine similarity by capturing structural and textural alignment that underlies patch transferability in QNNs.
    \item Our approach reduces attack success rate (ASR) by over 40\% on unseen patch and quantization configurations across CIFAR-10 and ImageNet, outperforming PBAT and DWQ with under 2\% drop in clean accuracy.
    \item Ablation studies validate the complementary role of each module and reveal that quantization alone does not sufficiently alter the shared attack surface—highlighting the need for targeted perceptual and structural misalignment.
\end{itemize}

\noindent In summary, \textbf{TriQDef} challenges the assumption that quantization inherently enhances adversarial robustness. By explicitly dismantling shared vulnerabilities at both the representational and gradient levels, TriQDef provides a principled and extensible framework for securing QNNs against patch-based threats.

\section{Related Work}
\label{sec:related}



Defensive strategies like Projected Gradient Descent (PGD)-based adversarial training~\cite{madry2017towards} offer limited effectiveness against patch attacks, which exploit model attention rather than gradient sensitivity. Patch-Based Adversarial Training (PBAT)~\cite{rao2020adversarial} incorporates patch patterns during training and improves robustness on seen configurations, but often fails to generalize across novel patch types or bit-widths. Other approaches, such as Double-Win Quantization~\cite{fu2021double}, stochastic precision inference~\cite{sen2020empir}, and feature-space smoothing~\cite{song2020improving}, primarily target pixel-level noise and do not directly tackle the structured, cross-bit nature of adversarial patches \cite{xiao2023robustmq}. 

\section{Motivation}
\label{sec:motiv}

Despite recent progress in adversarial training and quantization-aware techniques, we show that QNNs remain highly vulnerable to structured, localized adversarial attacks—particularly adversarial patches. This vulnerability stems from a critical oversight: existing defenses do not generalize across quantization levels, and thus fail to prevent cross-bit transferability of patch-based attacks. Our investigation reveals two key limitations that motivate the need for a principled, quantization-aware patch defense framework.

\noindent
\textbf{Adversarial Patches Transfer Effectively Across Bit-Widths.}
We begin by evaluating the transferability of adversarial patches crafted on full-precision (32-bit) models to quantized models trained using Quantization-Aware Training (QAT). Table~\ref{tab:experiment2_lavan} reports the Attack Success Rates (ASR) of two state-of-the-art patch attacks—LAVAN and GAP—on various QNN architectures. Notably, the adversarial patches retain high effectiveness even under extreme quantization (e.g., over 73\% ASR on 2-bit ResNet-56), confirming that quantization alone offers limited resilience against structured perturbations. This cross-bit vulnerability persists despite the reduced numerical precision and quantization noise introduced by QAT.

\begin{table}[ht]
    \centering
    \footnotesize
    \renewcommand{\arraystretch}{0.8}
    \setlength{\tabcolsep}{2.5pt}
    \begin{tabular}{c|c|c|c|c|c|c|c|c}
    \hline
    \textbf{Attack} & \multicolumn{4}{c|}{\textbf{LAVAN}} & \multicolumn{4}{c}{\textbf{GAP}} \\
    \hline
    \textbf{Model} & 32bit & 8bit & 4bit & 2bit & 32bit & 8bit & 4bit & 2bit \\
    \hline
    Res-56 & 86.43 & 83.24 & 76.22 & 73.08 & 84.40 & 56.69 & 54.22 & 47.91 \\
    Res-20 & 87.22 & 83.73 & 77.30 & 74.18 & 84.71 & 59.61 & 58.45 & 50.31 \\
    VGG-19 & 88.95 & 85.56 & 79.81 & 77.19 & 95.79 & 59.65 & 48.70 & 40.69 \\
    VGG-16 & 87.17 & 84.73 & 78.29 & 76.67 & 95.71 & 64.24 & 52.04 & 48.90 \\
    \hline
    \end{tabular}
    \caption{ASR (\%) of LAVAN and GAP (6x6 patches) transferred from full-precision models to QAT-trained QNNs on CIFAR-10.}
    \label{tab:experiment2_lavan}
\end{table}

\noindent
\textbf{Standard and Patch-Based Adversarial Training Fail to Generalize Across Bit-Widths.}
We further analyze whether existing adversarial defense methods can mitigate this vulnerability. In Table~\ref{tab:experiment4_patch_quant}, we compare standard adversarial training (AT) and Patch-Based Adversarial Training (PBAT) under different quantization paradigms: full-precision (FP), Quantization-Aware Training (QAT), and Post-Training Quantization (PTQ). While PBAT significantly reduces ASR for patch types seen during training, its robustness deteriorates sharply on unseen patch configurations—particularly when the patch was generated or tested under a different bit-width. For instance, ASR increases by more than 20\% when evaluated on 2-bit patches not seen during training. These results reveal a failure to generalize across quantization shifts, underscoring the need for defenses that explicitly target bit-level adversarial generalization.

\begin{table}[ht]
    \centering
    \footnotesize  
    \renewcommand{\arraystretch}{1.1}
    \setlength{\tabcolsep}{3pt}
    \begin{tabular}{l|ccc|ccc}
    \hline
    \textbf{Patch Type} & \multicolumn{3}{c|}{\textbf{Standard}} & \multicolumn{3}{c}{\textbf{PBAT}} \\
    \cline{2-7}
     & FP & QAT & PTQ & FP & QAT & PTQ \\
    \hline
    8$\times$8 (Seen)               & 88.17 & 81.56 & 85.24 & 40.39 & 40.56 & 45.44 \\
    10$\times$10 (Seen)             & 92.33 & 84.33 & 87.48 & 57.86 & 56.77 & 60.60 \\
    8$\times$8 (Unseen, 4-bit)      & 89.92 & 83.40 & 86.78 & 62.10 & 71.42 & 75.16 \\
    10$\times$10 (Unseen, 2-bit)    & 91.18 & 85.62 & 87.91 & 65.30 & 78.34 & 81.09 \\
    \hline
    \end{tabular}
    \caption{ASR (\%) of LAVAN attack across training paradigms and quantization levels. PBAT-trained models fail to generalize to unseen patch bit-widths.}
    \label{tab:experiment4_patch_quant}
\end{table}


\noindent
To further support these observations, we present in the Appendix: 
(1) cross-architecture transfer results including vision transformers; 
(2) analysis under dynamic and post-training quantization; 
(3) results using additional patch-based attacks (e.g., DPR, PatchAttack); 
and (4) comprehensive ablations and visualizations. These collectively highlight the limitations of existing defenses and motivate the design of \textbf{TriQDef}—a tri-level framework that breaks patch transferability via semantic, gradient, and curriculum-based alignment disruption.

\section{Methodology}
\label{sec:method}

\subsection{Overview}

We propose \textbf{TriQDef}, a unified defense framework that mitigates the transferability of patch-based adversarial attacks in QNNs. TriQDef integrates three complementary components into a cohesive training strategy. The first component, \textit{Feature Disalignment Penalty (FDP)}, disrupts semantic consistency by encouraging divergence in internal feature representations across different quantization levels. The second component, \textit{Gradient Perceptual Dissonance Penalty (GPDP)}, penalizes perceptual alignment in input gradients between bit-widths, targeting edge- and texture-level similarity that facilitates cross-bit transferability. The third component, \textit{Bit-Width-Aware Curriculum Training (BACT)}, stabilizes training under extreme quantization by progressively introducing lower-bit models, starting from a high-precision baseline. 

\subsection{Feature Disalignment Penalty (FDP)}
\label{sec:fdp}

As we show in Section~\ref{sec:motiv}, adversarial patches remain highly transferable across quantized neural network (QNN) variants, despite the reduced numerical precision. We argue that this transferability is facilitated by a phenomenon we call \textit{semantic alignment across bit-widths}—where internal representations across quantized models remain perceptually similar even under adversarial attack.

To assess this, we quantify perceptual similarity using two established descriptors: (i) \textbf{edge-based overlap}, computed using the Sobel operator and Intersection-over-Union (IoU)~\cite{zhang2018unreasonable}, and (ii) \textbf{textural similarity}, captured using Histogram of Oriented Gradients (HOG)~\cite{dalal2005histograms}, a robust descriptor widely used in computer vision and feature analysis.

\begin{figure}
    \centering
    \includegraphics[width=1\linewidth]{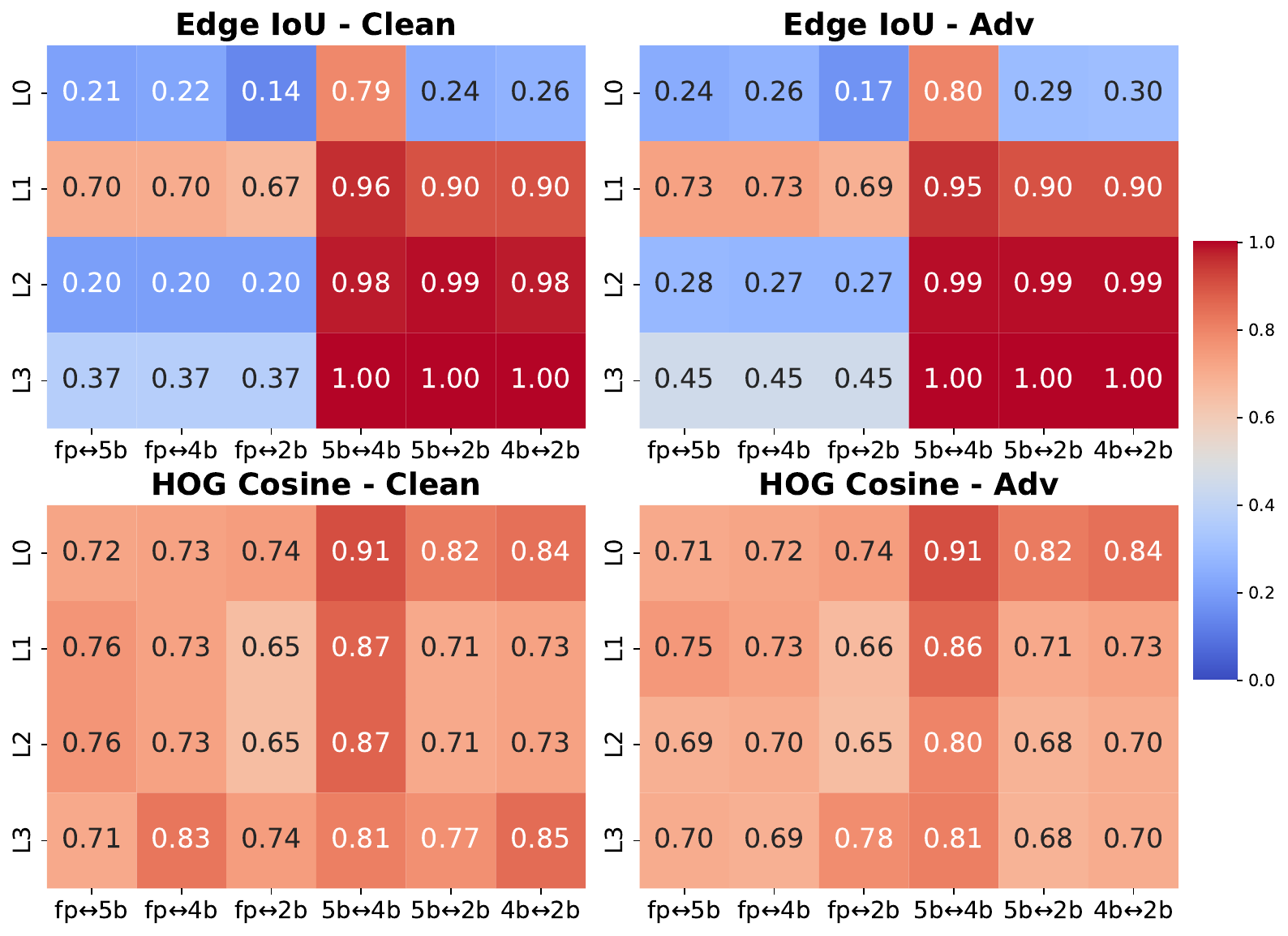}
    \caption{\textbf{Perceptual Alignment Across Bit-Widths.} 
    Heatmaps show pairwise \textbf{Edge IoU} and \textbf{HOG Cosine Similarity} across feature maps extracted at four convolutional layers in (L0–L3) from ResNet variants (fp, 5bit, 4bit, 2bit). High similarity persists even under adversarial input, enabling patch transferability.}
    \label{fig:similarity}
\end{figure}

Figure~\ref{fig:similarity} illustrates this behavior. Using a ResNet model on ImageNet, we analyze feature maps from clean and patched inputs across different bit-widths (full precision (fp), 5bit, 4bit, 2bit) and multiple layers. We compute pairwise Edge IoU and HOG Cosine Similarity as perceptual proxies. Results averaged over 100 samples reveal consistently high similarity, especially between adjacent quantization levels (e.g., 5bit $\leftrightarrow$ 4bit), indicating strong structural alignment that supports cross-bit patch generalization.

To mitigate this, we introduce the \textbf{Feature Disalignment Penalty (FDP)}—a regularizer inspired by perceptual similarity metrics and soft alignment losses~\cite{zhang2018unreasonable}. Unlike defenses that learn patch-specific filters, FDP explicitly discourages feature alignment across quantized variants by penalizing structural and textural similarity at selected intermediate layers during training.

Let \( f_b^{(l)}(x_{\text{adv}}) \) denote the activation at layer \( l \in \mathcal{L} \) of model \( f_b \) with bit-width \( b \), given a patched input \( x_{\text{adv}} \). FDP measures both structural and textural alignment between models \( b_i \ne b_j \) using two perceptual metrics:
\begin{itemize}
    \item Edge IoU: Intersection-over-Union between binarized Sobel edge maps of \( f_b^{(l)} \)
    \item HOG Cosine Similarity: Cosine similarity between Histogram of Oriented Gradient (HOG) descriptors
\end{itemize}

The FDP loss is defined as:
{\small
\begin{align}
\mathcal{L}_{\text{FDP}} = \sum_{l \in \mathcal{L}} \sum_{\substack{b_i, b_j \in \mathcal{B} \\ b_i \ne b_j}} 
\Big[ &\; \alpha \cdot \text{IoU} \big( \mathcal{E}(f_{b_i}^{(l)}(x_{\text{adv}})), \mathcal{E}(f_{b_j}^{(l)}(x_{\text{adv}})) \big) \nonumber \\
&+ \beta \cdot \cos \big( \phi(f_{b_i}^{(l)}(x_{\text{adv}})), \phi(f_{b_j}^{(l)}(x_{\text{adv}})) \big) \Big]
\end{align}
}

Because traditional perceptual metrics (e.g., hard thresholded edges, non-differentiable HOG) are not suitable for gradient-based optimization, we adapt recent differentiable approximations such as SoftDice~\cite{sudre2017generalised} and smooth HOG descriptors~\cite{KachouaneSLO15} to create an end-to-end trainable loss.

{\tiny
\begin{align}
\mathcal{L}_{\text{FDP}} = \sum_{l \in \mathcal{L}} \sum_{\substack{b_i, b_j \in \mathcal{B} \\ b_i \ne b_j}} 
\Big[ 
& \alpha \cdot \text{SoftDice}\left( S\left(E\left(f_{b_i}^{(l)}(x_{\text{adv}})\right) \right), S\left(E\left(f_{b_j}^{(l)}(x_{\text{adv}})\right) \right) \right) \nonumber \\
  & + \beta \cdot \cos\left( H\left(f_{b_i}^{(l)}(x_{\text{adv}})\right), H\left(f_{b_j}^{(l)}(x_{\text{adv}})\right) \right) 
\Big]
\end{align}}

\noindent where:
\begin{itemize}
    \item $E(\cdot)$: Computes Sobel edge magnitudes over spatial dimensions.
    \item $S(\cdot)$: Applies soft binarization using a sigmoid with quantile-based threshold: 
    \[
    S(A; \tau, k) = \sigma\left(k \cdot (A - \tau)\right), \quad \tau = \text{quantile}(A, q)
    \]
    with sharpness $k=100$, percentile $q=85$.
    \item $\text{SoftDice}(A, B) = \frac{2 \cdot \sum A \cdot B}{\sum A + \sum B + \epsilon}$.
    \item $H(\cdot)$: Computes a normalized HOG descriptor with $4\times4$ pixels per cell and $2\times2$ cells per block.
    \item $\alpha=0.5$, $\beta=1.0$: Weighting hyperparameters.
\end{itemize}

The choice of hyperparameters is based on ablation studies on their sensitivity (see Appendix \ref{sec:supp_ablation}).

\noindent\textbf{Why Use HOG Cosine Similarity?}  
HOG captures local edge orientation distributions and is robust to minor quantization noise and scale distortions. We observe high HOG similarity across bit-widths—indicating preserved structure—even when raw cosine similarity or Edge IoU degrade. This makes HOG a strong candidate for identifying transferable perceptual patterns.

\noindent\textbf{Why Use Edge-based SoftDice?}  
Edge IoU offers structural insight, but hard binarization is non-differentiable. SoftDice over softly binarized edges ensures smooth gradients while preserving interpretability. It captures shape-level alignment that HOG may overlook.

\noindent\textbf{Why Not LPIPS?} LPIPS, while widely used for perceptual similarity, is less suitable in our setting for several reasons. It is optimized for high-level semantic similarity based on human visual perception, making it less sensitive to the structural and directional patterns found in gradient maps or early-layer features that underlie patch transferability in quantized models. Moreover, LPIPS requires three-channel, large-resolution inputs and cannot be directly applied to single-channel or low-resolution feature maps or gradients. In contrast, Edge IoU and HOG Cosine Similarity offer lightweight, interpretable, and structurally grounded measures. They effectively capture spatial alignment (edges) and texture/orientation similarity (HOG) in both features and gradients, making them more appropriate for quantifying and disrupting perceptual alignment across bit-widths.

\noindent\textbf{Why FDP Works.}  
By introducing bit-level perceptual disalignment at key layers, FDP disrupts shared internal cues that adversarial patches exploit. This effectively breaks the cross-bit representational invariance that makes patches transferable.
This perceptual disalignment strategy complements adversarial training by targeting a root enabler of transferability—shared structure across models—which remains underexplored in the literature on quantized model robustness~\cite{li2024investigating}.
We provide a theoretical motivation for the FDP in Appendix~\ref{appendix:fdp_theory}, grounding its design in principles of perceptual alignment, representation similarity, and adversarial vulnerability across quantized models.

\noindent\textbf{Training with FDP.} 
Algorithm~\ref{alg:fdp} outlines our training procedure with the Feature Disalignment Penalty. For each batch, we apply an adversarial patch and extract intermediate features from multiple quantized models at selected layers. We then compute pairwise perceptual similarity using SoftDice over edge maps and cosine similarity over differentiable HOG descriptors. The total loss combines the standard cross-entropy with the FDP regularizer, guiding the models to develop divergent internal representations and weaken cross-bit adversarial transferability.

\begin{algorithm}
\caption{Compact FDP Training with Soft Disalignment}
\label{alg:fdp}
\KwIn{Quantized models $\{f_b\}$, adversarial patch $P$, batch $(x, y)$, target layers $\mathcal{L}$, mask $M$} 
\textbf{Hyperparameters:} $\lambda_{\text{FDP}}, \alpha, \beta$ \\
$x_{\text{adv}} \leftarrow x \odot (1 - M) + P \odot M$ \\
\ForEach{$b_i \in \mathcal{B}$,\ $l \in \mathcal{L}$}{
  $f_{b_i}^{(l)} \leftarrow f_{b_i}^{(l)}(x_{\text{adv}})$
}
$\mathcal{L}_{\text{FDP}} \gets 0$ \\
\ForEach{$l \in \mathcal{L}$,\ $(b_i, b_j), i \ne j$}{
  $E_i \leftarrow \text{Sobel}(\text{mean}(f_{b_i}^{(l)}))$,\quad
  $E_j \leftarrow \text{Sobel}(\text{mean}(f_{b_j}^{(l)}))$ \\
  $H_i \leftarrow \text{SoftHOG}(f_{b_i}^{(l)})$,\quad
  $H_j \leftarrow \text{SoftHOG}(f_{b_j}^{(l)})$ \\
  $\mathcal{L}_{\text{FDP}} \mathrel{+}= \alpha \cdot \text{SoftDice}(E_i, E_j) + \beta \cdot \cos(H_i, H_j)$
}
$\mathcal{L}_{\text{clean}} \gets \sum_{b} \mathcal{L}_{\text{CE}}(f_b(x), y)$,\quad
$\mathcal{L}_{\text{total}} \gets \mathcal{L}_{\text{clean}} + \lambda_{\text{FDP}} \cdot \mathcal{L}_{\text{FDP}}$ \\
Update all $f_b$ using $\mathcal{L}_{\text{total}}$
\end{algorithm}

\subsection{Gradient Perceptual Dissonance Penalty (GPDP)}
\label{sec:gpdp}

As shown in Table~\ref{tab:grad_similarity}, although gradient \textit{cosine similarity (CS)} is low across quantized models---suggesting directional disagreement---we observe persistently high perceptual similarity in the gradient maps. In particular, gradients exhibit strong \textit{HOG Cosine Similarity} and moderate \textit{Edge IoU}, revealing a hidden perceptual alignment that traditional CS fails to capture. We posit that this alignment facilitates the transferability of adversarial patches between bit-width variants by preserving texture and edge structure in gradient saliency.

\begin{table}[ht]
\centering
\footnotesize
\renewcommand{\arraystretch}{0.8}
\setlength{\tabcolsep}{2pt}
\begin{tabular}{ccccccc}
\toprule
\textbf{Metric} & \textbf{fp$\leftrightarrow$5b} & \textbf{fp$\leftrightarrow$4b} & \textbf{fp$\leftrightarrow$2b} & \textbf{5b$\leftrightarrow$4b} & \textbf{5b$\leftrightarrow$2b} & \textbf{4b$\leftrightarrow$2b} \\
\midrule
Cosine Sim. & 0.05 & 0.06 & 0.05 & 0.25 & 0.10 & 0.13 \\
Edge IoU & 0.14 & 0.15 & 0.14 & 0.20 & 0.15 & 0.15 \\
HOG CS & 0.81 & 0.81 & 0.80 & 0.82 & 0.81 & 0.81 \\
\bottomrule
\end{tabular}
\caption{Gradient similarity across bit-width models using different metrics. Despite low cosine similarity, perceptual metrics (HOG Cosine and Edge IoU) reveal strong structural alignment.}
\label{tab:grad_similarity}
\end{table}

To address this, we propose the \textbf{Gradient Perceptual Dissonance Penalty (GPDP)}—a perceptual regularizer designed to break gradient alignment across quantized models. GPDP penalizes both structural (edge-based) and textural (orientation-based) similarity in gradients, promoting gradient diversity that weakens the transferability of patch-based attacks.

Let $\nabla_{x}^{b_i}$ be the input gradient from a model quantized to bit-width $b_i$. We define the GPDP loss as:

{\small
\begin{align}
\mathcal{L}_{\text{GPDP}} = \sum_{\substack{b_i, b_j \in \mathcal{B} \\ b_i \ne b_j}} 
\Big[ &\; \alpha \cdot \text{SoftDice}\big( \text{Sobel}(\nabla_{x}^{b_i}), \text{Sobel}(\nabla_{x}^{b_j}) \big) \nonumber \\ & + \beta \cdot \cos\big( \text{SoftHOG}(\nabla_{x}^{b_i}), \text{SoftHOG}(\nabla_{x}^{b_j}) \big) \Big]
\end{align}}

Here, $\text{Sobel}(\cdot)$ computes edge maps from the gradient, and $\text{SoftHOG}(\cdot)$ is a differentiable version of the Histogram of Oriented Gradients (HOG) descriptor. SoftDice measures structural overlap, while the cosine of SoftHOG descriptors captures perceptual similarity. Coefficients $\alpha$ and $\beta$ balance these two components, with values set to $\alpha = 0.5$ and $\beta = 1.0$ in our experiments.

\noindent\textbf{Why GPDP Works.} Prior work \cite{yang2024quantization, tramer2018ensemble} has shown that adversarial transferability is tightly linked to gradient alignment. However, our findings indicate that even when gradients are directionally divergent (low cosine similarity), transfer persists due to structural similarity. GPDP directly penalizes this perceptual consensus, targeting especially early-layer gradient representations where saliency is concentrated. By diversifying gradient structure, GPDP reduces shared adversarial vulnerabilities across bit-widths.
We theoretically justify GPDP in Appendix~\ref{appendix:gpdp_theory}, showing that perceptual alignment in gradient structure—not just cosine similarity—enables patch transferability across bit-widths, and disrupting this alignment significantly weakens cross-bit attacks.
\begin{algorithm}
\caption{Training with GPDP}
\label{alg:gpdp}
\KwIn{Quantized models $\{f_b\}_{b \in \mathcal{B}}$, input batch $(x, y)$, adversarial version $x_{\text{adv}}$}
\textbf{Hyperparameters:} $\lambda_{\text{GPDP}}, \alpha, \beta$ \\
Initialize: $\mathcal{L}_{\text{GPDP}} \gets 0$ \\
\ForEach{$(b_i, b_j) \in \mathcal{B} \times \mathcal{B},\ i \ne j$}{
    $g_i \gets \nabla_x \mathcal{L}_{\text{CE}}(f_{b_i}(x_{\text{adv}}), y)$ \\
    $g_j \gets \nabla_x \mathcal{L}_{\text{CE}}(f_{b_j}(x_{\text{adv}}), y)$ \\
    $E_i \gets \text{Sobel}(g_i)$,\quad $E_j \gets \text{Sobel}(g_j)$ \\
    $H_i \gets \text{SoftHOG}(g_i)$,\quad $H_j \gets \text{SoftHOG}(g_j)$ \\
    $\mathcal{L}_{\text{GPDP}} \mathrel{+}= \alpha \cdot \text{SoftDice}(E_i, E_j) + \beta \cdot \cos(H_i, H_j)$
}
$\mathcal{L}_{\text{clean}} \gets \sum_{b \in \mathcal{B}} \mathcal{L}_{\text{CE}}(f_b(x), y)$ \\
$\mathcal{L}_{\text{total}} \gets \mathcal{L}_{\text{clean}} + \lambda_{\text{GPDP}} \cdot \mathcal{L}_{\text{GPDP}}$ \\
Update all $f_b$ using $\mathcal{L}_{\text{total}}$
\end{algorithm}
\noindent\textbf{Training with GPDP.}  
Algorithm~\ref{alg:gpdp} describes how the Gradient Perceptual Dissonance Penalty is integrated into the training loop. For each pair of quantized models, we compute the input gradients and apply perceptual similarity losses—based on edge structure and HOG texture—to penalize alignment. These losses are aggregated and added to the clean classification loss for joint optimization, thereby enforcing perceptual dissonance in gradient signals across bit-widths.
We apply GPDP \textit{only to adversarial inputs} to avoid impacting clean accuracy. It complements the Feature Disalignment Penalty (FDP) by targeting the gradient-level alignment that FDP cannot capture. The total training loss becomes:

\begin{equation}
\mathcal{L}_{\text{total}} = \mathcal{L}_{\text{clean}} + \lambda_{\text{FDP}} \cdot \mathcal{L}_{\text{FDP}} + \lambda_{\text{GPDP}} \cdot \mathcal{L}_{\text{GPDP}}
\end{equation}

\noindent where $\lambda_{\text{GPDP}}$ controls the strength of the regularization. We use $\lambda_{\text{GPDP}} = 0.5$ in practice.

\subsection{Bit-Width-Aware Curriculum Training (BACT)}
\label{sec:joint}

TriQDef unifies three complementary strategies—FDP, GPDP, and Bit-Width-Aware Curriculum Training (BACT)—into a cohesive framework to mitigate patch transferability in quantized neural networks (QNNs).

Training a joint ensemble of models across heterogeneous bit-widths (e.g., 32-bit to 2-bit) introduces convergence challenges due to severe quantization-induced distortions, gradient instability, and limited expressivity at low precision. To address this, we introduce \textbf{Bit-Width-Aware Curriculum Training (BACT)}, a staged training schedule that gradually incorporates lower bit-width models over time inspired by \cite{li2019additive}.

Directly training all bit-width variants jointly from the outset—especially including ultra-low-bit (e.g., 2-bit) models—leads to optimization interference and underfitting. Early exposure to unstable quantization noise hinders robust representation learning. BACT circumvents this by first training high-precision models (e.g., full-precision and 8-bit) to learn stable representations, then incrementally adding lower bit-width models (e.g., 5-bit, 4-bit, 2-bit) into the training loop. This progression improves generalization across the bit spectrum and stabilizes learning dynamics.

Given a predefined total number of epochs $T$, the training proceeds in phases. Each phase introduces a new bit-width $b_i \in \mathcal{B}$ according to a schedule (e.g., linear or staircase). For instance, a 4-stage schedule may start with $\{32\text{b}, 8\text{b}\}$ in phase 1, and add $\{5\text{b}, 4\text{b}, 2\text{b}\}$ in subsequent phases.

\noindent\textbf{Quantization-Aware Training.} All models in the curriculum are trained using Quantization-Aware Training (QAT)~\cite{li2019additive}, which simulates quantization during both forward and backward passes. This enables each model to adapt to the precision constraints during training, unlike Post-Training Quantization (PTQ) that applies quantization after training. QAT improves the robustness and accuracy of quantized models under adversarial conditions.
To further improve stability, we initialize each low-bit model from the weights of the previously trained higher-bit model (e.g., initialize 4-bit model from the 5-bit checkpoint). This warm-start reduces gradient shocks and accelerates adaptation to lower precision constraints.

BACT (i) improves convergence by isolating unstable gradients in early stages, (ii) reduces representational interference between precision levels, and (iii) enhances overall robustness to adversarial patches by allowing each quantized model to learn under controlled, progressively challenging conditions.


\noindent\textbf{Runtime Efficiency.}
While TriQDef performs multi-bit training during optimization, inference requires only a single forward pass under the target bit-width. This ensures zero inference-time overhead, full compatibility with integer-only deployment, and robust performance across a wide range of quantization levels.

\section{Results and Analysis}
\label{results}

\subsection{Experimental Setup}

\noindent\textbf{Datasets.}  
We evaluate our proposed methods on two widely used benchmark datasets: CIFAR-10~\cite{cifar} and ImageNet~\cite{krizhevsky2017imagenet}.

\noindent\textbf{Model Architectures.}  
Our experiments cover a broad spectrum of architectures:  
ResNet-56, ResNet-34, ResNet-20, and ResNet-18~\cite{he2016deep}, VGG-16 and VGG-19~\cite{simonyan2014very}, AlexNet~\cite{krizhevsky2017imagenet}, Inception-v3~\cite{szegedy2016rethinking}, DenseNet-121~\cite{huang2018dense}. Swin-S~\cite{liu2021swin} and DeiT-B~\cite{deit} (used only to evaluate patch transferability).

\noindent\textbf{Patch-Based Attacks.}  
To assess vulnerability to structured adversarial perturbations, we evaluate against several state-of-the-art patch-based attacks:
LAVAN~\cite{lavan}, Adversarial Patch (GoogleAP)~\cite{googleap}, Deformable Patch Representation (DPR)~\cite{chen2022shape}, and the black-box PatchAttack~\cite{yang2020patchattack}.

\noindent\textbf{Implementation Details.}  
All experiments are conducted using PyTorch on NVIDIA V100 GPUs. We use a batch size of 128 and train models with SGD (momentum 0.9, weight decay $1\times10^{-4}$). The learning rate is initialized to 0.1 and decayed by a factor of 10 at 50\% and 75\% of training. Models are trained for 200 epochs on CIFAR-10 and 120 epochs on ImageNet.
For adversarial training, we generate a diverse offline pool of adversarial patches using the full-precision model. These patches vary in size and position to improve robustness against diverse patch-based threats.

\noindent\textbf{Quantization Setup.}  
We adopt QAT~\cite{li2019additive} with simulated quantization applied via the straight-through estimator (STE). Both weights and activations are quantized using symmetric uniform quantization with per-tensor scale and zero-point. The set of target bit-widths is $\mathcal{B} = \{2, 4, 5, 32\}$, and training is done via progressive introduction through BACT.

\noindent\textbf{FDP and GPDP Hyperparameters.}  
The regularization coefficients are set as follows: $\lambda_{\text{FDP}} = 0.8$, and $\lambda_{\text{GPDP}} = 0.5$. The used hyperparameters were selected based on ablation studies on their sensitivity (see Appendix \ref{sec:supp_ablation} for more details).

\subsection{Clean Accuracy Under TriQDef Across Bit-Widths}

A critical goal of TriQDef is to improve robustness without sacrificing clean accuracy. Table~\ref{tab:qadt_r_clean_acc_combined} shows the clean performance of models trained with TriQDef compared to Standard QAT and PBAT across multiple bit-widths. While adversarial training introduces a slight accuracy drop, TriQDef maintains competitive performance—closely matching QAT and outperforming PBAT across both CIFAR-10 and ImageNet. Notably, TriQDef avoids the overfitting and degradation seen in PBAT at lower bit-widths, demonstrating its effectiveness in preserving model expressiveness while enhancing robustness.

\begin{table}[ht]
    \centering
    \footnotesize  
    \renewcommand{\arraystretch}{1.0}
    \setlength{\tabcolsep}{5pt}
    \begin{tabular}{c|c|c|c|c|c}
        \hline
        \multirow{2}{*}{\textbf{Defense}} & \multirow{2}{*}{\textbf{Dataset}} & \multicolumn{4}{c}{\textbf{Clean Accuracy (\%)}} \\
        \cline{3-6}
        & & \textbf{32bit} & \textbf{5bit}  & \textbf{4bit} & \textbf{2bit} \\
        \hline
        Standard QAT & CIFAR-10  & 89.4 & 85.1  & 80.5 & 78.2 \\
        PBAT         & CIFAR-10  & 88.2 & 81.6  & 77.8 & 75.5 \\
        \textbf{TriQDef (Ours)} & CIFAR-10 & 89.4  & 83.3 & 78.2 & 75.8 \\
        \hline
        Standard QAT & ImageNet  & 85.2 & 79.3  & 77.5 & 73.9 \\
        PBAT         & ImageNet  & 84.1 & 77.3  & 74.2 & 71.8 \\
        \textbf{TriQDef (Ours)} & ImageNet  & 85.2 & 78.1 & 75.1 & 72.5 \\
        \hline
    \end{tabular}
    \caption{Clean accuracy (\%) of ResNet-56 (CIFAR-10) and ResNet-34 (ImageNet) trained under Standard QAT, PBAT, and TriQDef across various bit-widths.}
    \label{tab:qadt_r_clean_acc_combined}
\end{table}

\subsection{Effect of Bit-Width on Adversarial Robustness}
We evaluate TriQDef's robustness across multiple quantization levels (32bit, 5bit, 4bit, 2bit) using two patch-based attacks: LAVAN and GAP. Results are compared against PBAT and DWQ on both seen and unseen patch configurations. Generalization is evaluated using unseen patches, defined as patches not encountered during training. The reported ASR reflects an average across diverse attacks involving variations in patch size, spatial placement, and generation bit-width.

\begin{table}[ht]
    \centering
    \footnotesize
    \renewcommand{\arraystretch}{1}
    \setlength{\tabcolsep}{5pt}
    \begin{tabular}{c|c|c|c|c|c}
        \hline
        \textbf{Defense} & \textbf{Dataset} & \textbf{32b} & \textbf{5b} & \textbf{4b} & \textbf{2b} \\
        \hline
        PBAT & CIFAR-10 & 51.4 & 46.7 & 43.2 & 39.7 \\
        DWQ  & CIFAR-10 & 87.9 & 82.4 & 77.2 & 76.4 \\
        \textbf{TriQDef} & CIFAR-10 & \textbf{32.4}  & \textbf{30.2}  & \textbf{28.4} & \textbf{26.2} \\
        PBAT (Unseen) & CIFAR-10 & 77.2 & 73.7 & 67.8 & 65.3 \\
        DWQ (Unseen)  & CIFAR-10 & 87.8 & 82.6 & 77.9 & 76.3 \\
        \textbf{TriQDef (Unseen)} & CIFAR-10 & \textbf{35.6}  & \textbf{33.3} & \textbf{29.1}  & \textbf{27.3}\\
        \hline
        PBAT & ImageNet & 53.4 & 49.2 & 41.8 & 37.1 \\
        DWQ  & ImageNet & 86.7 & 71.3 & 63.5 & 61.2 \\
        \textbf{TriQDef} & ImageNet & \textbf{35.0} &  \textbf{33.1} & \textbf{31.1}  & \textbf{28.5} \\
        PBAT (Unseen) & ImageNet & 78.5 & 73.7 & 67.2 & 64.6 \\
        DWQ (Unseen)  & ImageNet & 86.9 & 72.3 & 62.4 & 60.2 \\
        \textbf{TriQDef (Unseen)} & ImageNet & \textbf{37.1} & \textbf{35.3} & \textbf{32.5} & \textbf{30.7} \\
        \hline
    \end{tabular}
    \caption{ASR (\%) under LAVAN attack (6×6 patches) on CIFAR-10 and (64x64 patches)  on ImageNet across bit-widths and patch generalization settings. Lower is better.}
    \label{tab:triqdef_lavan}
\end{table}

\begin{table}
    \centering
    \footnotesize
    \renewcommand{\arraystretch}{1}
    \setlength{\tabcolsep}{5pt}
    \begin{tabular}{c|c|c|c|c|c}
        \hline
        \textbf{Defense} & \textbf{Dataset} & \textbf{32b} & \textbf{5b} & \textbf{4b} & \textbf{2b} \\
        \hline
        PBAT & CIFAR-10 & 48.7 & 43.5 & 40.1 & 37.9 \\
        DWQ  & CIFAR-10 & 85.4 & 80.9 & 75.6 & 73.5 \\
        \textbf{TriQDef} & CIFAR-10 & \textbf{29.7} & \textbf{22.1} & \textbf{19.3} & \textbf{17.2} \\
        PBAT (Unseen) & CIFAR-10 & 75.3 & 71.2 & 66.9 & 63.2 \\
        DWQ (Unseen)  & CIFAR-10 & 86.1 & 80.3 & 75.2 & 72.3 \\
        \textbf{TriQDef (Unseen)} & CIFAR-10 &  \textbf{32.5}  & \textbf{30.8}  & \textbf{27.3} & \textbf{25.5}\\
        \hline
        PBAT & ImageNet & 50.2 & 46.1 & 39.8 & 35.4 \\
        DWQ  & ImageNet & 85.3 & 70.5 & 61.4 & 59.5 \\
        \textbf{TriQDef} & ImageNet & \textbf{33.2} & \textbf{28.1} & \textbf{26.3} & \textbf{23.1} \\
        PBAT (Unseen) & ImageNet & 76.5 & 71.3 & 65.7 & 62.1 \\
        DWQ (Unseen)  & ImageNet & 84.3 & 70.1 & 61.7 & 58.2 \\
        \textbf{TriQDef (Unseen)} & ImageNet & \textbf{39.4} & \textbf{37.3}  & \textbf{34.5} & \textbf{32.9} \\
        \hline
    \end{tabular}
    \caption{ASR (\%) under GAP attack on CIFAR-10 and ImageNet across bit-widths and patch generalization settings. Lower is better.}
    \label{tab:triqdef_gap}
\end{table}

\noindent TriQDef consistently achieves the lowest ASR across all bit-widths and attacks, including 2-bit models, where it surpasses PBAT and DWQ by over 20\% and 50\%, respectively. On unseen patches, TriQDef maintains strong generalization with only a 1.6–2.3\% ASR increase, compared to 25–27\% for PBAT, confirming its resistance to patch overfitting. DWQ offers negligible robustness, further highlighting the effectiveness of TriQDef in countering transferable patch threats under extreme quantization.
Additional experiments involving alternative architectures and diverse attack methods are provided in Appendix~\ref{sec:supp_results}.

\subsection{TriQDef vs. Inference-Time Preprocessing Defenses}
Existing inference-time preprocessing defenses, such as Jedi~\cite{tarchoun2023jedi}, are tailored for full-precision models and rely on high-resolution entropy maps and intermediate features to localize and inpaint adversarial patches. However, these methods face critical limitations in the quantized setting: reduced bit precision in QNNs severely degrades feature granularity and dynamic range, making entropy-based localization unreliable. Moreover, Jedi’s inpainting modules (e.g., autoencoders) introduce floating-point dependencies and computational overhead incompatible with the low-latency, integer-only constraints of edge deployments. DiffPure~\cite{nie2022DiffPure}, a purification-based defense using score-based diffusion models, is even more impractical in this context, requiring 5.58–17.14 seconds per ImageNet image and over 7\,GB of GPU memory for inference.
In contrast, \textbf{TriQDef} is a training-time-only defense with no inference-time overhead, offering full compatibility with quantized and resource-constrained environments. On ImageNet, TriQDef significantly outperforms Jedi in robust accuracy across multiple bit-widths—even when Jedi is allowed to retain full-precision preprocessing—highlighting TriQDef’s superior generalization and practicality under extreme quantization constraints.

\begin{table}[ht]
\centering
\footnotesize
\renewcommand{\arraystretch}{0.9}
\setlength{\tabcolsep}{6pt}
\begin{tabular}{c|c|c|c}
\hline
\textbf{Defense Method} & \textbf{32bit (FP)} & \textbf{5bit QNN} & \textbf{2bit QNN} \\
\hline
Jedi & 64.3 & 52.1 & 23.4 \\
\textbf{TriQDef (Ours)} & \textbf{78.3} & \textbf{76.1} & \textbf{65.8} \\
\hline
\end{tabular}
\caption{Robust accuracy (\%) under LAVAN attack on ImageNet (ResNet-50) comparing Jedi and TriQDef across quantization levels. The higher the better.}
\label{tab:jedi_vs_triQDef}
\end{table}

\subsection{Impact of TriQDef Components}

To evaluate the contribution of each component in \textbf{TriQDef}, we conduct an ablation study by selectively disabling its core modules. 

\noindent\textbf{Effect of Removing FDP:}  
Excluding the FDP module leads to a sharp rise in vulnerability, with ASR exceeding 55\% on CIFAR-10 and over 52\% on ImageNet at 2-bit precision. This indicates that representational alignment across bit-widths enables strong cross-bit transferability. Without FDP, the model fails to decouple semantic encoding across quantization configurations, resulting in high susceptibility to both seen and unseen adversarial patches.

\noindent\textbf{Effect of Removing GPDP:}  
Disabling GPDP results in a consistent increase in ASR across all bit-widths. Since GPDP disrupts gradient alignment across quantized variants, its absence allows adversarial gradients to converge more effectively during attack optimization. The ASR increases by more than 10\% compared to the full TriQDef setup, highlighting the importance of controlling cross-bit optimization consistency.

\noindent\textbf{Full TriQDef:}  
The complete TriQDef framework achieves the lowest ASR across all configurations. It exhibits strong generalization to \emph{unseen} patch patterns with only marginal performance degradation compared to seen attacks. This demonstrates the complementary nature of FDP and GPDP in severing both semantic and gradient-based pathways for adversarial transferability.


\begin{table}[ht]
    \centering
    \footnotesize
    \renewcommand{\arraystretch}{1}
    \setlength{\tabcolsep}{5pt}
    \begin{tabular}{c|c|c|c|c|c}
    \hline
    \textbf{Defense Config} & \textbf{Seen/Unseen} & \textbf{32bit} & \textbf{5bit}  & \textbf{4bit} & \textbf{2bit} \\ \hline
    w/o FDP & Seen   & 65.2 & 61.1  & 59.7 & 55.9    \\ 
    w/o GPDP  & Seen   & 43.2   & 41.7 & 39.4 &  37.6   \\ 
    \textbf{TriQDef (Full)} & Seen & \textbf{32.4}  & \textbf{30.2}  & \textbf{28.4} & \textbf{26.2} \\ \hline
    w/o FDP & Unseen & 65.2 & 61.1  & 59.7 & 55.9   \\ 
    w/o GPDP  & Unseen & 48.8  & 46.9   & 43.7 & 41.8    \\ 
    \textbf{TriQDef (Full)} & Unseen & \textbf{35.6}  & \textbf{33.3} & \textbf{29.1}  & \textbf{27.3}\\ \hline
    \end{tabular}
    \caption{Ablation study: ASR (\%) of LAVAN (6×6 patches) attack across bit-widths on CIFAR-10 (ResNet-56) under seen and unseen patch settings.}
    \label{tab:ablation_combined_cifar}
\end{table}

\begin{table}[ht]
    \centering
    \footnotesize
    \renewcommand{\arraystretch}{1}
    \setlength{\tabcolsep}{5pt}
    \begin{tabular}{c|c|c|c|c|c}
    \hline
    \textbf{Defense Config} & \textbf{Seen/Unseen} & \textbf{32bit} & \textbf{5bit}  & \textbf{4bit} & \textbf{2bit} \\ \hline
    w/o FDP & Seen   & 68.7 & 60.2  & 55.3 & 52.1   \\ 
    w/o GPDP  & Seen   & 48.6 & 46.8 & 44.2 &   42.5   \\ 
    \textbf{TriQDef (Full)} & Seen & \textbf{35.0} &  \textbf{33.1} & \textbf{31.1}  & \textbf{28.5}\\ \hline
    w/o FDP & Unseen & 68.7 & 60.2  & 55.3 & 52.1   \\ 
    w/o GPDP  & Unseen & 42.6  & 40.8  & 37.5 & 35.6    \\ 
    \textbf{TriQDef (Full)} & Unseen & \textbf{37.1} & \textbf{35.3} & \textbf{32.5} & \textbf{30.7}\\ \hline
    \end{tabular}
    \caption{Ablation study: ASR (\%) of LAVAN attack (64x64) across bit-widths on ImageNet (ResNet-34) under seen and unseen patch settings.}
    \label{tab:ablation_combined_imagenet}
\end{table}

\section{Conclusion}
We presented \textbf{TriQDef}, a principled defense framework aimed at mitigating the transferability of patch-based adversarial attacks in QNNs. TriQDef combines three synergistic components—FDP, GPDP, and BACT—to explicitly dismantle semantic and gradient-level alignment across bit-widths. Unlike prior defenses that rely on adversarial patch augmentation, TriQDef targets the root cause of patch transferability by disrupting both feature- and gradient-level consensus among quantized models. Our experiments show that TriQDef significantly reduces attack success rates across unseen patches and bit-width combinations, while preserving clean accuracy and avoiding inference-time overhead. 


{
    \small
    \bibliographystyle{ieeenat_fullname}
    \bibliography{main}
}

\clearpage
\setcounter{page}{1}
\maketitlesupplementary


\section{Extended Patch Transferability Results}
\label{app:motivation}


\subsection{Cross-Architecture Transfer Evaluation}
Patch-based adversarial attacks not only persist across different quantization bit-widths but also exhibit strong cross-architecture transferability, making them a severe security threat in real-world black-box settings. To evaluate this phenomenon, we generate adversarial patches on a base architecture (e.g., ResNet-20 at 32-bit precision) and transfer them to models with different architectures (e.g., ResNet-56, VGG-16, and VGG-19) trained with QAT at various bit-widths (8-bit, 5-bit, 4-bit, and 2-bit). The Attack Success Rate is recorded for each architecture-bitwidth combination to assess the transferability of adversarial patches across both architectural and quantization changes.

\begin{table}[ht]
    \centering
    \footnotesize  
    \renewcommand{\arraystretch}{0.8}  
    \setlength{\tabcolsep}{2.5pt}  
    \begin{tabular}{|c|c|c|c|c|c|c|c|c|c}
    \hline
          &   \multicolumn{4}{c|}{ \textbf{ResNet56}} &  \multicolumn{4}{c|}{\textbf{VGG-19}}  \\ \hline
     Bitwidth      & 32bit & 8bit & 4bit & 2bit & 32bit & 8bit & 4bit & 2bit \\
    \hline                 
      \textbf{ResNet20} &     84.17 & 79.62  & 77.66 & 75.21  &  78.82 & 74.53 & 72.15 & 70.21 \\
    \hline
          & \multicolumn{4}{c|}{ \textbf{ResNet20} }& \multicolumn{4}{c|}{  \textbf{VGG-19}}  \\ \hline
     Bitwidth      & 32bit & 8bit & 4bit & 2bit & 32bit & 8bit & 4bit & 2bit \\
          \hline      
      \textbf{ResNet56} &  84.11  & 77.67  & 75.33 & 71.76  & 77.43 & 75.09  & 73.82 & 71.22\\
    \hline
           & \multicolumn{4}{c|}{  \textbf{ResNet20} }& \multicolumn{4}{c|}{  \textbf{VGG-16}} \\ \hline
     Bitwidth      & 32bit & 8bit & 4bit & 2bit & 32bit & 8bit & 4bit & 2bit \\
          \hline     
      \textbf{VGG-19} &  83.23  & 80.87  & 78.11 &  75.32 & 85.32 & 80.42  & 78.23 & 76.44\\
    \hline
            & \multicolumn{4}{c|}{ \textbf{VGG-19}} &  \multicolumn{4}{c|}{  \textbf{ResNet56} } \\ \hline
     Bitwidth     & 32bit & 8bit & 4bit & 2bit & 32bit & 8bit & 4bit & 2bit \\
          \hline     
      \textbf{VGG-16}  & 83.29 & 78.48 & 76.65 &  74.39 & 80.55 &  78.93  & 75.34  & 73.87 \\
    \hline
    \end{tabular}
    \caption{ASR (\%) transfer across different QNNs with different bitwidths and architectures on CIFAR-10. }
    \label{tab:experimet5}
\end{table}

As presented in Table~\ref{tab:experimet5}, a patch generated on ResNet-20 achieves an 84.17\% attack success rate on 32-bit ResNet-56 and 78.82\% on 32-bit VGG-19. Patches created on VGG-19 and VGG-16 maintain high success rates when tested on ResNet architectures, confirming their strong cross-architecture transferability. Even at low-bit settings (e.g., 2-bit), patches retain attack success rates above 70\%, highlighting their resilience under quantization-induced transformations.

\subsection{Patch Transferability under Post-Training Quantization (PTQ)}

We next evaluate patch transfer to PTQ models on ImageNet using ResNet-18 and ResNet-34. Table~\ref{tab:experiment2_lavan_resnet34} shows that even at 2-bit precision, patches maintain over 50\% ASR. These findings confirm that bit-depth reduction—even without adversarial training—does not inherently block patch effectiveness.

\begin{table}[ht]
    \centering
    \footnotesize
    \renewcommand{\arraystretch}{0.9}
    \setlength{\tabcolsep}{3pt}
    \begin{tabular}{c|c|c|c|c||c|c|c|c}
    \hline
    & \multicolumn{4}{c|}{\textbf{ResNet-34}} & \multicolumn{4}{c}{\textbf{ResNet-18}} \\
    \cline{2-9}
    \textbf{NP} & 32bit & 5bit & 4bit & 2bit & 32bit & 5bit & 4bit & 2bit \\
    \hline
    0.10 & 99.31 & 66.32 & 63.56 & 56.31 & 99.98 & 72.63 & 67.89 & 65.76 \\
    0.08 & 98.08 & 64.91 & 59.97 & 52.25 & 99.93 & 66.37 & 61.11 & 55.42 \\
    0.06 & 97.12 & 64.79 & 57.31 & 50.43 & 96.01 & 58.20 & 53.59 & 51.84 \\
    \hline
    \end{tabular}
    \caption{ASR (\%) of LAVAN patches under PTQ on ImageNet ResNet models.}
    \label{tab:experiment2_lavan_resnet34}
\end{table}



\subsection{Transformer Architectures Are Equally Susceptible}

\noindent Table~\ref{tab:experiment2_transformer} presents the vulnerability of transformer-based models—specifically Swin-S and DeiT-B—quantized using post-training quantization (PTQ) techniques. We evaluate both MinMax and Percentile calibration methods under two patch-based attacks: LAVAN and GAP. Despite being structurally distinct from convolutional architectures, these models remain highly susceptible to adversarial patches, with attack success rates (ASR) exceeding 60\% even in their 8-bit quantized forms.
These results underscore the generality of patch-based threats across architectural paradigms. The persistence of high ASR across different calibration methods and both attacks suggests that transformer quantization does not inherently mitigate adversarial vulnerability, reinforcing the necessity of robust, architecture-agnostic defenses such as TriQDef.

\begin{table}[ht]
    \centering
    \footnotesize  
    \renewcommand{\arraystretch}{0.9} 
    \setlength{\tabcolsep}{3pt} 
    \begin{tabular}{c|c|c|c|c|c}
    \hline
    & \textbf{Attack} & \multicolumn{2}{c|}{\textbf{LAVAN}} & \multicolumn{2}{c}{\textbf{GAP}} \\ 
    \hline
    \textbf{Model} & \textbf{Calibration} & \textbf{32-bit} & \textbf{8-bit} & \textbf{32-bit} & \textbf{8-bit} \\
    \hline
    Swin-S & MinMax     & 91.80 & 62.11 & 85.32 & 59.84 \\
    Swin-S & Percentile & 93.10 & 63.72 & 87.19 & 61.23 \\
    DeiT-B & MinMax     & 93.63 & 64.76 & 88.03 & 62.44 \\
    DeiT-B & Percentile & 90.12 & 61.51 & 84.37 & 58.73 \\
    \hline
    \end{tabular}
    \caption{ASR (\%) of LAVAN and GAP (64×64 patches) on PTQ Swin-S and DeiT-B evaluated on ImageNet. Results are shown for MinMax and Percentile calibration methods.}
    \label{tab:experiment2_transformer}
\end{table}


\subsection{Patch Transferability under Dynamic Quantization (DQ)}

\noindent To evaluate the effectiveness of adversarial patches under more flexible deployment settings, we assess attack success rates (ASR) on dynamically quantized (DQ) 8-bit models. Unlike static quantization, DQ applies quantization to weights at runtime, commonly used for latency-efficient inference on general-purpose CPUs.

\noindent Table~\ref{tab:dyn_ptq} presents ASR results for LAVAN and GAP attacks on various CIFAR-10 models, comparing full-precision (32-bit) and dynamically quantized 8-bit versions. The results demonstrate that patch-based attacks retain high transferability and effectiveness, even under dynamic quantization schemes. Notably, models like ResNet-20 and VGG variants still suffer from ASR values exceeding 70\% in many cases, with minimal degradation compared to their 32-bit counterparts.
These findings emphasize that dynamic quantization alone is insufficient to mitigate the threat of physical adversarial patches. Thus, defenses like TriQDef remain essential even in low-bit dynamic settings.

\begin{table}[ht]
    \centering
    \footnotesize
    \renewcommand{\arraystretch}{0.7}
    \setlength{\tabcolsep}{3.5pt}
    \begin{tabular}{c|c|c|c|c|c|c|c|c}
        \hline
        \textbf{Model} & \multicolumn{2}{c|}{\textbf{ResNet-56}} & \multicolumn{2}{c|}{\textbf{ResNet-20}} & \multicolumn{2}{c|}{\textbf{VGG-19}} & \multicolumn{2}{c}{\textbf{VGG-16}} \\
        \hline
        \textbf{Bit} & 32-bit & 8-bit & 32-bit & 8-bit & 32-bit & 8-bit & 32-bit & 8-bit \\
        \hline
        LAVAN & 86.43 & 84.03 & 87.22 & 83.29 & 88.95 & 76.33 & 87.17 & 71.58 \\
        GAP   & 84.40 & 82.40 & 84.71 & 53.76 & 95.71 & 54.12 & 95.79 & 41.78 \\
        \hline
    \end{tabular}
    \caption{ASR (\%) of LAVAN and GAP attacks (6×6 patches) across dynamically quantized 8-bit models on CIFAR-10. Despite runtime quantization, adversarial patches maintain high transferability.}
    \label{tab:dyn_ptq}
\end{table}

\section{Theoretical Justification}
In this section, we provide a theoretical justification for FDP and GPDP. 
\subsection{Theoretical Justification for Feature Disalignment Penalty (FDP)}
\label{appendix:fdp_theory}

FDP is grounded in theoretical principles from adversarial robustness, representation learning, and gradient alignment. It is designed to break a key enabler of patch-based attack transferability: the \emph{semantic alignment} of internal representations across quantized models.

\paragraph{Transferability via Representation Alignment.} 
Let $f_b$ denote a model quantized to bit-width $b \in \mathcal{B}$, and let $f_b^{(l)}(x)$ denote its activation at layer $l$. Let $x_{\text{adv}}$ be an adversarially patched input crafted to fool a surrogate model $f_{b_i}$. The patch transfers successfully to a target model $f_{b_j}$ if:
\[
f_{b_i}^{(l)}(x_{\text{adv}}) \approx f_{b_j}^{(l)}(x_{\text{adv}}) \quad \Rightarrow \quad f_{b_i}(x_{\text{adv}}) \approx f_{b_j}(x_{\text{adv}}),
\]
i.e., shared internal features lead to similar high-level decisions. Thus, representational alignment is a \textit{sufficient condition} for adversarial patch transfer. FDP aims to break this alignment by minimizing:
\[
\mathcal{L}_{\text{FDP}} \propto \sum_{l} \sum_{b_i \ne b_j} \text{Sim}(f_{b_i}^{(l)}(x_{\text{adv}}), f_{b_j}^{(l)}(x_{\text{adv}})),
\]
which encourages divergence of internal features across bit-widths, especially for adversarial inputs.

\paragraph{Representation Learning Perspective.}
From the perspective of representation learning, FDP functions similarly to a contrastive loss. By penalizing similarity between features of different models on the same input, it promotes feature \textit{decorrelation} across quantized variants. This aligns with findings from contrastive learning~\cite{wang2020understanding} and ensemble robustness~\cite{fort2020deep}, where diversity in intermediate representations improves generalization and robustness.

\paragraph{Gradient-Based Justification.}
FDP also implicitly induces \textit{gradient disalignment}. Since input gradients are a function of intermediate features (via backpropagation), dissimilarity in internal activations leads to divergence in $\nabla_x \mathcal{L}(f_b(x), y)$. This weakens the ability of a patch optimized on $f_{b_i}$ to be effective on $f_{b_j}$:
\[
f_{b_i}^{(l)}(x_{\text{adv}}) \not\approx f_{b_j}^{(l)}(x_{\text{adv}}) \quad \Rightarrow \quad \nabla_x \mathcal{L}(f_{b_i}) \not\approx \nabla_x \mathcal{L}(f_{b_j}),
\]
thus reducing gradient-based attack transferability.

\paragraph{Saliency and Interpretability Alignment.}
Prior work~\cite{zhang2018unreasonable,hooker2019benchmark} suggests that robust models exhibit unique and spatially localized saliency patterns. By minimizing perceptual similarity across feature maps (e.g., via HOG and edge-based metrics), FDP reduces the spatial overlap of vulnerable regions across quantized models. This discourages universal patch activation across the bit spectrum.

In summary, FDP is theoretically justified because it:
\begin{itemize}
    \item Breaks the sufficient condition of cross-model feature alignment.
    \item Encourages bit-specific feature specialization via a contrastive-like loss.
    \item Induces input gradient divergence across bit-widths.
    \item Prevents shared saliency patterns, lowering cross-bit patch vulnerability.
\end{itemize}
These principles collectively reduce adversarial patch transferability across quantized neural networks.


\subsection{Theoretical Justification of GPDP}
\label{appendix:gpdp_theory}

The effectiveness of adversarial examples is largely attributed to the alignment of gradient directions across models~\cite{tramer2017space, wang2021unified}. In the case of quantized neural networks (QNNs), despite differences in numerical precision, adversarial perturbations often transfer between bit-widths because the input gradients of different QNNs remain structurally and perceptually similar—even when their cosine similarity is low (see Table~\ref{tab:grad_similarity}). This perceptual alignment enables an adversarial patch optimized on one quantized model to activate similar vulnerable patterns in another.

Let $\nabla_x^{b_i}$ denote the gradient of a quantized model with bit-width $b_i$ with respect to input $x$, and let $\mathcal{A}_{\text{adv}}(x) = x + \delta$ denote an adversarial transformation computed using gradient ascent:
\[
\delta = \epsilon \cdot \text{sign}(\nabla_x^{b_i} \mathcal{L}(f_{b_i}(x), y))
\]
The success of $\delta$ on a different model $f_{b_j}$ depends on the local alignment between $\nabla_x^{b_i}$ and $\nabla_x^{b_j}$~\cite{ilyas2019adversarial}. While cosine similarity measures vector alignment, it fails to capture local structural and textural similarities that are critical for patch-based attacks, which rely on spatially localized perturbations.

We define the following perceptual similarity-based decomposition of transferability:
\begin{align*}
\mathcal{T}(b_i \rightarrow b_j) \propto\ 
&\underbrace{\cos\left(\nabla_x^{b_i},\ \nabla_x^{b_j}\right)}_{\text{directional}} \\
&+ \underbrace{\text{EdgeIoU}\left(\nabla_x^{b_i},\ \nabla_x^{b_j}\right)}_{\text{spatial structure}} \\
&+ \underbrace{\cos\left(\text{HOG}(\nabla_x^{b_i}),\ \text{HOG}(\nabla_x^{b_j})\right)}_{\text{textural similarity}}
\end{align*}

This shows that transferability arises not only from vector similarity but also from *perceptual consensus* in gradient maps. Thus, to reduce cross-bit adversarial success, we must disrupt both the directional and perceptual agreement in gradients.

The \textbf{Gradient Perceptual Dissonance Penalty (GPDP)} does precisely this by penalizing:
\begin{itemize}
    \item Structural similarity via differentiable Edge IoU between edge maps of $\nabla_x^{b_i}$ and $\nabla_x^{b_j}$.
    \item Textural similarity via cosine similarity between soft HOG descriptors of gradients.
\end{itemize}

By introducing gradient-level dissonance across QNNs, GPDP increases the difficulty of crafting perturbations that remain effective across models, thus mitigating cross-bit transferability. This aligns with theoretical findings in~\cite{tramer2017space, ilyas2019adversarial} that successful transferability relies on shared gradient-based decision boundaries.

Therefore, GPDP is a principled regularizer that enforces \textit{gradient-space fragmentation}, complementing FDP's \textit{feature-space disalignment} to build a more comprehensive defense.

\section{Additional Ablation Studies}
\label{sec:supp_ablation}

\subsection{Hard vs. Soft Perceptual Metrics}
\label{sec:hard_vs_soft}

To validate our choice of perceptual alignment losses used in \textbf{FDP} and \textbf{GPDP}, we compare the behavior of hard metrics (non-differentiable) such as Edge Intersection-over-Union (Edge IoU) and HOG Cosine Similarity with their soft, differentiable counterparts: SoftDice and SoftHOG Cosine. The goal is to measure structural similarity between feature maps and gradients across models quantized to different bit-widths.

\noindent\textbf{Hard Metrics.}
As shown in Table~\ref{tab:hard_soft_metrics}, Edge IoU and HOG Cosine reveal significant perceptual alignment between bit-width variants, especially for nearby pairs such as 5bit $\leftrightarrow$ 4bit. For instance, in layer \texttt{L3.conv1}, all intra-quantized model pairs yield an Edge IoU of 1.0—indicating perfect edge alignment—while HOG similarities frequently exceed 0.8. However, such saturation diminishes their utility for gradient-based optimization and weakens their discriminative power, particularly in deeper layers.

\noindent\textbf{Soft Metrics.}
In contrast, SoftDice and SoftHOG produce a smoother, more nuanced similarity landscape across both shallow and deep layers. For example, in \texttt{L0.conv1}, SoftDice similarity between int5 $\leftrightarrow$ 4bit is 0.86, while the cross-bit pair fp $\leftrightarrow$ 2bit yields a significantly lower score of 0.50. This dynamic range allows us to effectively penalize both low-frequency and high-frequency structural similarities in the loss function. Moreover, unlike hard metrics, soft variants avoid saturation and remain responsive throughout training, making them highly suitable for alignment regularization.

\noindent\textbf{Justification for Loss Design.}
These results support our design choice to adopt \textbf{SoftDice and SoftHOG} in both FDP and GPDP. They provide differentiable approximations of perceptual similarity while capturing critical edge and texture-level redundancies across quantized models—precisely the structural alignments that enable patch transferability.

\begin{table}[ht]
\centering
\footnotesize
\renewcommand{\arraystretch}{1.1}
\setlength{\tabcolsep}{5pt}
\begin{tabular}{|c|c|c|c|c|}
\hline
\textbf{Metric} & \textbf{Pair} & \textbf{L0.conv1} & \textbf{L1.conv1} & \textbf{L2.conv1} \\
\hline
Edge IoU & fp $\leftrightarrow$ 5bit & 0.2464 & 0.7367 & 0.2802 \\
SoftDice & fp $\leftrightarrow$ 5bit & 0.4492 & 0.2697 & 0.6214 \\
\hline
HOG Cosine & fp $\leftrightarrow$ 5bit & 0.7177 & 0.7536 & 0.6900 \\
SoftHOG & fp $\leftrightarrow$ 5bit & 0.7214 & 0.7614 & 0.7116 \\
\hline
Edge IoU & int5 $\leftrightarrow$ 4bit & 0.8094 & 0.9576 & 0.9872 \\
SoftDice & int5 $\leftrightarrow$ 4bit & 0.8638 & 0.7776 & 0.8634 \\
\hline
HOG Cosine & int5 $\leftrightarrow$ 4bit & 0.9145 & 0.8664 & 0.8004 \\
SoftHOG & int5 $\leftrightarrow$ 4bit & 0.9163 & 0.8709 & 0.7931 \\
\hline
\end{tabular}
\caption{Comparison of Hard (Edge IoU, HOG Cosine) vs. Soft (SoftDice, SoftHOG) metrics between bit-width variants in early layers. Shown: similarity scores for selected pairs in \texttt{conv1}.}
\label{tab:hard_soft_metrics}
\end{table}

\subsection{Sensitivity to Loss Hyperparameters}
\label{sec:loss_ablation}

To evaluate the sensitivity of TriQDef to its loss hyperparameters, we conduct an ablation study by varying the weights associated with its two main components: the \textbf{Feature Disalignment Penalty (FDP)} and the \textbf{Gradient Perceptual Dissonance Penalty (GPDP)}. Specifically, we analyze the impact of scaling coefficients $(\alpha, \beta)$ for bit-aware patch training and $(\lambda_{\text{FDP}}, \lambda_{\text{GPDP}})$ for perceptual alignment disruption across multiple quantization levels.

We report the clean accuracy and adversarial robustness (ASR \%) under the LAVAN attack (6×6 patches) on CIFAR-10 across 32-bit, 5-bit, 4-bit, and 2-bit models. The setting Clean refers to clean input evaluation, while Adv denotes adversarial inputs.

\begin{table}[ht]
    \centering
    \footnotesize
    \renewcommand{\arraystretch}{1.1}
    \setlength{\tabcolsep}{5pt}
    \begin{tabular}{|c|c|c|c|c|c|c|}
        \hline
        \textbf{Param.} & \textbf{Values} & \textbf{Setting} & \textbf{32b} & \textbf{5b} & \textbf{4b} & \textbf{2b} \\
        \hline
        ($\alpha$,$\beta$) & (1.0,1.0) & Clean & 82.1 & 75.4 & 71.2 & 68.3 \\
        ($\alpha$,$\beta$) & (1.0,1.0) & Adv   & 53.8 & 51.4 & 50.6 & 49.1 \\
        \hline
        ($\alpha$,$\beta$) & (0.5,1.0) & Clean & 84.2 & 80.3 & 79.9 & 77.8 \\
        ($\alpha$,$\beta$) & (0.5,1.0) & Adv   & 50.0 & 47.5 & 45.2 & 42.1 \\
        \hline
        ($\alpha$,$\beta$) & \textbf{(1.0,0.5)} & Clean & 85.2 & 78.1 & 75.1 & 72.5 \\
        ($\alpha$,$\beta$) & \textbf{(1.0,0.5)} & Adv  & 54.86 & 52.0 & 53.2 & 52.7 \\
        \hline
        ($\lambda_{\text{FDP}}$,$\lambda_{\text{GPDP}}$) & (1.0 , 1.0) & Clean & 80.7 & 69.3 & 64.8 & 61.5 \\
        ($\lambda_{\text{FDP}}$,$\lambda_{\text{GPDP}}$) & (1.0 , 1.0) & Adv   & 42.1 & 40.2 & 40.1 & 39.0 \\
        \hline
        ($\lambda_{\text{FDP}}$,$\lambda_{\text{GPDP}}$) & (0.5 , 0.8) & Clean & 87.6 & 80.5 & 76.1 & 73.4 \\
        ($\lambda_{\text{FDP}}$,$\lambda_{\text{GPDP}}$) & (0.5 , 0.8) & Adv   & 50.8 & 43.7 & 41.3 & 40.2 \\
        \hline
        ($\lambda_{\text{FDP}}$,$\lambda_{\text{GPDP}}$) & \textbf{(0.8 , 0.5)} & Clean & 85.2 & 78.1 & 75.1 & 72.5 \\
        ($\lambda_{\text{FDP}}$,$\lambda_{\text{GPDP}}$) & \textbf{(0.8 , 0.5)} & Adv   & 54.86 & 52.0 & 53.2 & 52.7 \\
        \hline
    \end{tabular}
    \caption{Average Model accuracy (\%) under clean and adversarial settings (LAVAN 6×6 patch) on CIFAR-10 across bit-widths, varying alignment and patch generation loss coefficients. Higher accuracy are preferable.}
    \label{tab:ablation}
\end{table}


Table~\ref{tab:ablation} presents a detailed ablation study analyzing the impact of the patch generation losses $(\alpha, \beta)$ and the alignment regularization weights $(\lambda_{\text{FDP}}, \lambda_{\text{GPDP}})$ on clean and adversarial accuracy across different quantization levels on CIFAR-10.

\begin{itemize}
    \item \textbf{Patch Loss Weights $(\alpha, \beta)$:} The configuration $(1.0, 1.0)$ offers moderate clean accuracy but exhibits reduced robustness under attack (e.g., 53.8\% at 32-bit). Lowering $\beta$ to $0.5$—as in $(1.0, 0.5)$—improves both clean and adversarial accuracy across bit-widths. This suggests that deemphasizing bit-width-specific loss during patch generation helps create perturbations that generalize better across quantized models. Conversely, the setting $(0.5, 1.0)$ yields the highest clean accuracy (up to 79.9\% at 4-bit), but at the cost of significant robustness degradation, indicating a trade-off between clean accuracy and adversarial resistance.

    \item \textbf{Disalignment Loss Weights $(\lambda_{\text{FDP}}, \lambda_{\text{GPDP}})$:} Strong penalties (e.g., $(1.0, 1.0)$) reduce both clean and adversarial performance, likely due to training instability or over-regularization. Moderate weights such as $(0.5, 0.8)$ enhance clean accuracy and slightly improve robustness. The configuration $(0.8, 0.5)$ emerges as the most balanced setting, offering strong clean accuracy and the lowest adversarial degradation (e.g., 54.86\% at 32-bit, 52.0\% at 5-bit), supporting its selection as the default configuration in TriQDef.

    \item \textbf{Consistency Across Bit-Widths:} The observed trends are consistent from 32-bit to 2-bit, demonstrating that TriQDef maintains its effectiveness even in extreme low-bit settings. This validates the bit-aware robustness and generalization capabilities of our framework.

\end{itemize}

\section{Additional Results}
\label{sec:supp_results}

\subsection{Results on Additional Architectures}

To demonstrate the generality of TriQDef, we evaluate its effectiveness across multiple network architectures on both CIFAR-10 and ImageNet. We report attack success rates (ASR \%) under the LAVAN patch-based attack across different quantization levels (32-bit to 2-bit). The results consistently show that TriQDef significantly reduces ASR, confirming its robustness across architectures and datasets (See Tables \ref{tab:appx_cifar} and \ref{tab:appx_imagenet}).


\begin{table}[ht]
    \centering
    \footnotesize  
    \renewcommand{\arraystretch}{0.95} 
    \setlength{\tabcolsep}{5pt} 
    \begin{tabular}{c|c|c|c|c|c}
        \toprule
        \textbf{Model} & \textbf{Setting} & \textbf{32bit} & \textbf{5bit} & \textbf{4bit} & \textbf{2bit} \\ \hline
         VGG-16        & No defense       & 87.17  & 81.45 & 78.29 & 76.67  \\ 
         VGG-16        & TriQDef          & 29.34  & 27.10 & 26.43 & 21.20  \\ \hline
         VGG-19        & No defense       & 88.95  & 82.28 & 79.81 & 77.19  \\   
         VGG-19        & TriQDef          & 28.70  & 25.20 & 22.30 & 20.90  \\ \hline  
         ResNet-20     & No defense       & 87.22  & 80.65 & 77.30 & 74.18  \\ 
         ResNet-20     & TriQDef          & 30.24  & 27.30 & 26.90 & 22.60  \\ 
        \bottomrule
    \end{tabular}
    \caption{ASR (\%) of LAVAN attack (6×6 patch) on CIFAR-10 across multiple architectures and quantization levels.}
    \label{tab:appx_cifar}
\end{table}


\begin{table}[ht]
    \centering
    \footnotesize  
    \renewcommand{\arraystretch}{0.95} 
    \setlength{\tabcolsep}{5pt} 
    \begin{tabular}{c|c|c|c|c|c}
        \toprule
        \textbf{Model} & \textbf{Setting} & \textbf{32bit} & \textbf{5bit} & \textbf{4bit} & \textbf{2bit} \\ \hline
         ResNet-18     & No defense       & 99.93  & 66.37 & 61.11 & 55.42  \\ 
         ResNet-18     & TriQDef          & 33.50  & 31.30 & 29.60 & 27.40  \\ \hline
         Inception v3  & No defense       & 89.10  & 57.21 & 55.32 & 50.66  \\   
         Inception v3  & TriQDef          & 35.60  & 32.10 & 30.40 & 27.30  \\ \hline  
         MobileNetV2   & No defense       & 88.35  & 59.43 & 54.97 & 49.25  \\ 
         MobileNetV2   & TriQDef          & 29.53  & 27.80 & 25.10 & 23.50  \\ 
        \bottomrule
    \end{tabular}
    \caption{ASR (\%) of LAVAN attack with 64×64 patch on ImageNet across architectures and quantization levels.}
    \label{tab:appx_imagenet}
\end{table}

\subsection{Results for Other Attacks}
\noindent\textbf{Results under DRP Attack.}

\noindent The DRP attack~\cite{chen2022shape} introduces shape-deformable adversarial patches that adaptively alter their structure and appearance to exploit neural network vulnerabilities. Unlike traditional pixel-level perturbations, DRP leverages geometric transformations to improve both robustness and transferability, making it particularly effective in black-box and cross-model scenarios.

We evaluate the robustness of TriQDef against DRP on both CIFAR-10 and ImageNet across multiple quantization levels. As shown in Table~\ref{tab:triqdef_drp}, TriQDef consistently outperforms prior defenses, including PBAT and DWQ, under both standard and unseen patch settings. Notably, TriQDef maintains a significant ASR reduction, even under the unseen patch regime where generalization is critical.

\begin{table}[ht]
    \centering
    \footnotesize
    \renewcommand{\arraystretch}{1}
    \setlength{\tabcolsep}{5pt}
    \begin{tabular}{c|c|c|c|c|c}
        \hline
        \textbf{Defense} & \textbf{Dataset} & \textbf{32b} & \textbf{5b} & \textbf{4b} & \textbf{2b} \\
        \hline
        PBAT & CIFAR-10 & 56.6 & 48.3 & 46.5 & 43.2 \\
        DWQ  & CIFAR-10 & 87.9 & 82.4 & 77.2 & 76.4 \\
        \textbf{TriQDef} & CIFAR-10 & \textbf{35.4}  & \textbf{31.7}  & \textbf{30.1} & \textbf{28.8} \\
        PBAT (Unseen) & CIFAR-10 & 81.4 & 75.4 & 71.8 & 68.4 \\
        DWQ (Unseen)  & CIFAR-10 & 90.2 & 84.3 & 80.6 & 78.3 \\
        \textbf{TriQDef (Unseen)} & CIFAR-10 & \textbf{42.7}  & \textbf{35.5} & \textbf{31.2}  & \textbf{29.6}\\
        \hline
        PBAT & ImageNet & 60.4 & 53.7 & 50.8 & 48.5 \\
        DWQ  & ImageNet & 88.6 & 80.4 & 75.3 & 71.6 \\
        \textbf{TriQDef} & ImageNet & \textbf{45.6} &  \textbf{40.7} & \textbf{38.1}  & \textbf{35.3} \\
        PBAT (Unseen) & ImageNet & 81.1 & 75.2 & 71.2 & 67.9 \\
        DWQ (Unseen)  & ImageNet & 91.9 & 80.3 & 75.4 & 70.2 \\
        \textbf{TriQDef (Unseen)} & ImageNet & \textbf{48.1} & \textbf{45.7} & \textbf{38.5} & \textbf{35.2} \\
        \hline
    \end{tabular}
    \caption{ASR (\%) under DRP attack (6×6 patches on CIFAR-10 and 64×64 patches on ImageNet) across bit-widths and generalization settings. Lower is better.}
    \label{tab:triqdef_drp}
\end{table}

\noindent\textbf{Results under PatchAttack.} PatchAttack~\cite{yang2020patchattack} is a reinforcement learning (RL)-based framework for black-box adversarial patch generation. It iteratively learns patch placement and content via model queries without requiring gradient access.
To evaluate the transferability of PatchAttack against quantized models, we adopted the original full-precision (32-bit) patch optimization setup while targeting 5-bit quantized ResNet-50 models. Specifically, the attack was crafted on the full-precision model and evaluated on both the vanilla and TriQDef-defended quantized models to assess robustness across bit-widths.

As shown in Table~\ref{tab:triqdef_patchattack}, the 5-bit model without defense exhibits a notable drop in ASR compared to the 32-bit baseline, yet remains vulnerable to transfer-based patch attacks. In contrast, our \textbf{TriQDef} framework significantly reduces the ASR on both 32-bit and 5-bit models, demonstrating its ability to disrupt patch transferability under query-efficient, black-box settings.

\begin{table}[ht]
    \centering
    \footnotesize
    \renewcommand{\arraystretch}{1.1}
    \setlength{\tabcolsep}{6pt}
    \begin{tabular}{c|c|c}
        \hline
        \textbf{Defense} & \textbf{32bit} & \textbf{5bit} \\
        \hline
        No Defense       & 90.23           & 75.42          \\
        \textbf{TriQDef} & \textbf{43.20}  & \textbf{25.40} \\
        \hline
    \end{tabular}
    \caption{Attack Success Rate (ASR \%) under PatchAttack on ResNet-50 (ImageNet) using a 6\% patch area and 9500 queries. Attacks are crafted on the 32-bit model and transferred to both full-precision and 5-bit variants.}
    \label{tab:triqdef_patchattack}
\end{table}

\end{document}